\RequirePackage{fix-cm}
\documentclass[twocolumn]{svjour3}          
\smartqed  
\usepackage{graphicx}
\usepackage{amsmath, amssymb}
\usepackage{float}
\usepackage{multicol}
\begin{document}
\title{Static object detection and segmentation in videos based on dual foregrounds difference with noise filtering}
\author{Waqqas-ur-Rehman Butt \and
       Martin Servin 
}
\institute{Waqqas-ur-Rehman Butt \and Martin Servin \at
              Umea University \at              
              90187, SE \at
              \email{ waqqas.butt@umu.se}}          
\date{Received: date / Accepted: date}
\maketitle

\begin{abstract}
This paper presents static object detection and segmentation method in videos from cluttered scenes. Robust static object detection is still challenging task due to presence of moving objects in many surveillance applications. The level of difficulty is extremely influenced by on how you label the object to be identified as static that do not establish the original background but appeared in the video at different time. In this context, background subtraction technique based on the frame difference concept is applied to the identification of static objects. Firstly, we estimate a frame differencing foreground mask image by computing the difference of each frame with respect to a static reference frame. The Mixture of Gaussian \emph{MOG} method is applied to detect the moving particles and then outcome foreground mask is subtracted from frame differencing foreground mask. Pre-processing techniques, illumination equalization and de-hazing methods are applied to handle low contrast and to reduce the noise from scattered materials in the air e.g. water droplets and dust particles. Finally, a set of mathematical morphological operation and largest connected-component analysis is applied to segment the object and suppress the noise. The proposed method was built for rock breaker station application and effectively validated with real, synthetic and two public data sets. The results demonstrate the proposed approach can robustly detect, segmented the static objects without any prior information of tracking.

\end{abstract}

\section{Introduction}
\label{intro}
Static object detection and segmentation is an exciting task in video processing. It deals with identifying an object of interest and then tracking it over time, including its presence, shape and position. It has attracted great attention in video surveillance applications such as illegal parked vehicle detection in traffic monitoring system and abandoned object detection on public areas. 


In the recent years, many approaches have been proposed with the aim to achieve the precise detection of moving objects in a robust and efficient way. Adaptive background subtraction technique used to separate moving object from the background \cite{Stauffer}. Some approaches are based on the combination of visual features statistical analysis and temporal analysis of motion, where visual features are segmented into different frames and integration with similar motion vectors. These approaches show significant accuracy in moving objects detection. However, complications arise due to different environmental conditions such as lightening, contrast and scattering particles. Particularly, moving shadows and poor contrast can affect the robust object detection. There are two main sources of information in the videos: visual features (colour,shape,texture), and motion information are used to identify and tracking the objects. The combination of these sources generally leads to approaches that are more robust. 

Static object detection and segmentation in videos is still problematic and remains open research task.A robust approach should be able fully detect the foreground objects in the videos.
In this context, we present an approach, aim to detect the static objects in videos. We particularly emphasis on how to detect the individual larger size of static object (rock or piles of rocks) in real and synthetic videos. The image sequence where the objects are in motion is assumed to be unusable. In addition, when the objects have relaxed to a static state the environmental conditions are harsh, with dust, water droplets and poor contrast. We apply two approaches simultaneously to detect and segment of static objects. Firstly, we applied the frame difference approach to estimate the difference between current and static reference frame. The foreground model using this approach is not precise due to the moving objects. Therefore, \emph{BS-MOG} \cite{Stauffer,Bouwmans08backgroundmodeling} applied to detect the moving particles. In the next, eliminate the moving pixels by subtracting the \emph{BS} foreground mask to frame difference foreground mask.  Furthermore, pre-processing methods are applied to reduce the effects from poor contrast and lightening. Finally, the post-processing methods applied to detect and segment the desired object.

The rest of this paper is organized as follows. Section 2 reviews the related work on object detection in videos including key challenges relevant for the application of rock piles. Section 3 describes the datasets achieved at the station and simulated videos. Section 4 briefly describes the proposed method including pre-processing and post-processing methods and segmentation of object. Section 5 shows the experimental results and implementation remarks. In the final section, concludes the paper with future work and references.
\section{Related Work}
\label{Related Work}
Background subtraction (BS) based techniques have become the popular choice due to the common use of static cameras in environments where the illumination changes in the scene are gradual and foreground objects are detected by comparing the current frame with background model \cite{Beynon2003,BS03,1400815}. The background model is built by first or previous frame. Finally, they update the background model. A static background hypothesis is usually used in background subtraction, which is not applicable in real environment. 
With indoor scenes, reflections on screens lead to background changes. Similarly, due to wind, rain or illumination changes brought by weather, static backgrounds methods have complications with outdoor scenes \cite{1233909}. Though, this approach is not used for analysing static objects \cite{1400815}. 
The background model must be frequently kept updated to avoid the influence of background variation. An accurate, reliable, and flexible background model is essential for robust tracking \cite{1233909}. There are several BS methods have been suggested in the last few decades including Running Gaussian Average \cite{598236} , Temporal Median Filter \cite{925356}, Mixture of Gaussians \cite{Stauffer,Bouwmans08backgroundmodeling} , Kernel Density Estimation (KDE) \cite{Elgammal:2000:NMB:645314.649432}, Kalman Filter \cite{411746}, and Co-occurrence of Image Variations \cite{1211453}. Therefore, BS techniques do not be enough for the static object detection and are thus enhanced by an additional approach. 

Static object detection is getting special consideration in crowded environments by considering appearance variations such as colour composition, shape, lighting changes in the scene \cite{Sengar2017MovingOD}. Tracking information of the object used in \cite{4425318,4425317,5279721}.These methods can find difficulties due to occlusions and shadows generated by moving objects, which turn the object initialization and tracking into a hard problem to solve.

In \cite{Porikli2007}, a pixel wise system approach proposed to address the limitations of tracking based approach by using dual foregrounds. In this approach, two background models, a short term and a long term, were used with different learning rate. In this way, the background models absorb the static objects and detect them as those sets of pixels classified as background by the short term model but not by the long term background model. Furthermore, after a given time depending on its learning rate the static objects may also absorb by the long-term model and system failed to detect those static objects. To reduce this problem, the background model is updated selectively. Another drawback of this approach is that improper updates give incorrect detection of the static object.  

The foreground masks obtained from the subtraction of two background models by selectively updating the short-term model used to precise segmentation of moving object detection in \cite{Elgammal:2000:NMB:645314.649432}, but they did not consider the problem of detecting new static objects. 

Static object detection by selectively updating the long-term models with dual background models and finite-state-machine is described in \cite{5279721,HerasEvangelio2010,Lan2016} . In \cite{HerasEvangelio2010} approach, the system is restricted to the detection of static objects, which do not belong to the scene background and actual scene background must be known when the state machine starts working, which is a realistic requirement. In \cite{Lan2016}, this approach is based on updating decisions and on tracking information.

A two-frame and three-frame differencing methods \cite{ZHANG20122705,FERRYMAN2013789,Sengar2017MovingOD} work well in moving regions but fail to detect the static object due to the dynamic changes in the background image and the quality of the foreground object is highly dependent on a fixed threshold. Furthermore, frame difference approach mentioned in \cite{Sengar2017MovingOD} is based on the simple \emph{OR} operation to get the foreground objects which does not suitable for noisy scenes.


Earlier methods are inadequate for complex scenes as many incorrect detection are produced due to the scattering mediums and high amount of detected foreground. The irradiance received by the camera from the scene point is attenuated along the line of sight by atmospheric particles \cite{5567108}. The amount of scattering materials depends on the distance of the scene points from the camera, the degradation is spatially variant. Different methods proposed in \cite{855874,1201821} to cope this problem. Some assume having multiple images of the same scene under different conditions. Recently, a method based on a single image has been proposed \cite{5567108}, which tackle the lack of constraints by incorporating various priors. In this paper, we propose static object detection method in videos with improving the efficiency and accuracy of the detection and segmentation.
\section{Data sets}
\label{sec:2}
The method is tested on four data sets:  real, synthetic and two publicly available data sets i-LIDS \cite{iLids} and CAVIAR \cite{CAVIAR}. The real data set come from a rock breaker station in an underground mine at Boliden Garpenberg with high-resolution $(1920 \times 1090)$ static camera with 29 FPS in different environmental conditions such as dust, water droplets, poor contrast and colour similarities.  It is impractical to show all the frames of the video so three frames are selected randomly as shown in Fig. 1(a). A truck enters in the station and drop rock material on the grid surface.
Rocks that are larger than the grid size will remain on the grid while the smaller sized rocks passes through and fall down. As per application point of view, detect the large rocks in order to steer the rock breaker tool to hammer them into smaller pieces. The secondary, synthetic, data set is made from a 3D-model of the rock breaker station constructed using SpaceClaim and rendered with photo-realistic refraction and shadows using Key-shot as  to investigate the dependency on camera angles, light variations, shape, position, size of rocks and moisture particles effects as shown in Fig. 1 (b).
\begin{figure}[!ht]
\centering
 \includegraphics[scale=.40]{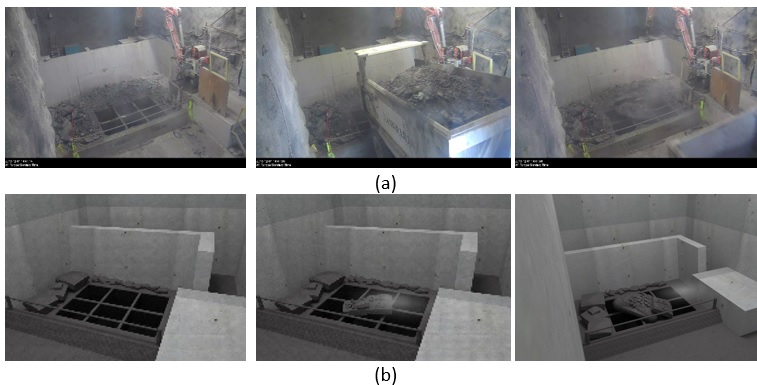}
\caption{Inner view of station:(a)Real video(b)Synthetic video}
\label{fig:1}       
\end{figure}
\section{Proposed Method}
\label{sec:2}
In this section, we propose a new method in three parts: Pre-Processing; Frame differencing and BS-MOG method and Post-Processing. Fig. 2 illustrate the method as a block diagram. Each part is designed as modular processing unit, making it easy to modify any part provided the input and output data types remain compatible with the connecting parts.
\begin{figure}[!ht]
  \includegraphics[scale=.65]{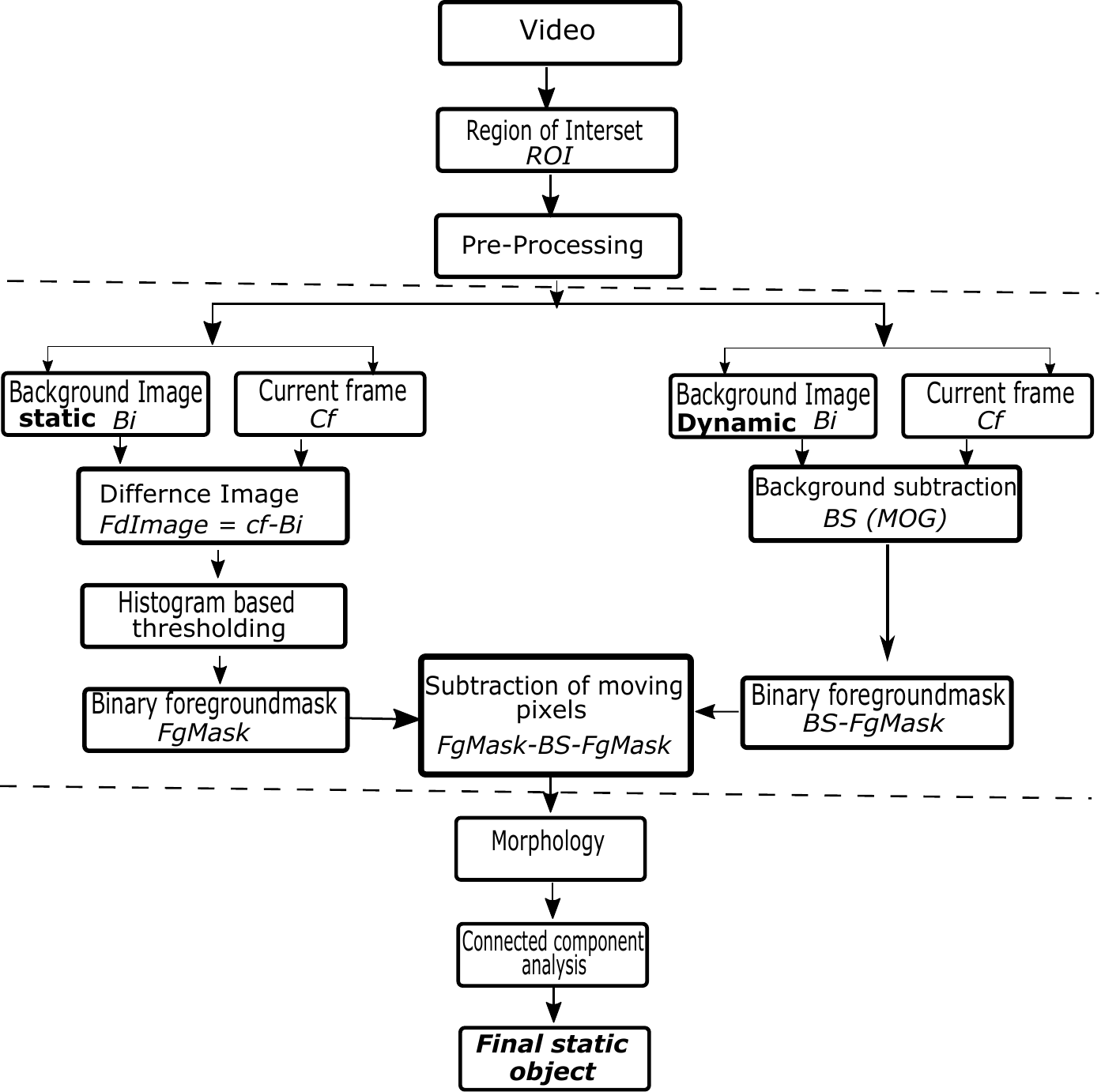}
\caption{Block diagram of overall system}
\label{fig:2}       
\end{figure}
\subsection{Pre-Processing}
\subsubsection{Region of Interest (ROI)}
\label{sec:2}
Initially, video is loaded from the data set and split into individual frame images.
Many methods are found in the literature \cite{Benrabha:2017:ARD:3109761.3158381,CHENG2010299} for detecting the region of interest \emph{ROI}. These methods are divided into two types; manual/semi-automatic and fully automated. In order to reduce the computational time, we used manual method to determine the \emph{ROI} by user interaction with the system. While there is no user intervention involved in fully automated method \cite{Lan2016}. A user define \emph{ROI} with $(300 \times 300)$ resolution is selected in the first frame where the rock material will be drop as shown in Fig. 3(c). 
\begin{figure}[H]
\centering
  \includegraphics[width=80mm]{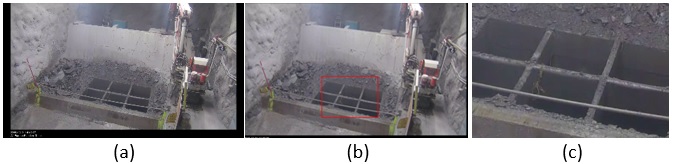}
\caption{Defining the region of interest: (a) Original frame, (b) Draw ROI, (c) ROI}
\label{fig:3}       
\end{figure}
\subsubsection{Illumination Equalization}
\label{sec:2}
Illumination equalization plays a significant role to reduce the noise when lightening is irregular. This noise is composed of pixels which are different in appearance with neighbouring pixels and can be suppressed by averaging in the similar area of true image data. Actually, true image data is able to share the similarities in these averaged areas but noise in these areas is not. Therefore, this method will hold true image data efficiently and reduce the noise. 
The method \cite{CHEUNG20123336} works exclusively on the luminance value of the individual pixel but has a few flaws such as the effects of irregular lighting are attenuated only along the single direction vertically. Moreover, fixed scaling size of image $(71\times144)$ and mask size $(3\times3)$ is used. We used the improved method presented in \cite{Butt2016,Butt2017} making it more robust respect to illumination effects which can adapt to the multiple directions of the lighting and local regions $(2p+1) \times (2q+1)$ as shown in Fig. 4(a).
\begin{figure}[h]
\begin{center}
\includegraphics[scale=2.2]{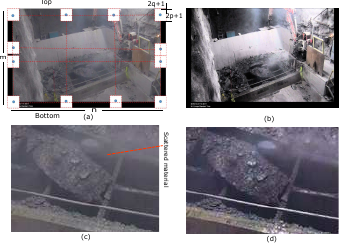}
\caption{Illumination equalization and De-Hazing:(a)original frame image (b)output image after illumination equalization (c)Water droplets and dust on rock(d)De-hazing function [23]}
\label{fig:4}       
\end{center}
\end{figure}
The \emph{ROI} image size $m \times n$ provided to the input of the function is initially converted in \texttt{HSV} colour space, let $L(i,j)$ and $L'(i,j)$ represent respectively the luminance of each pixel before and after the operation of equalization. Each pixel in ROI image assumes the value obtained from the calculation of the average of the luminance values of all the pixels included in the mask that flows throughout the image along the two main directions identified. The luminance value of the pixels obtained by the application formulated (Horizontal and Vertical) directions is described in formula \ref{Eq 1}  and \ref{Eq 2}.
\begin{equation}
L'(i,j)
= L(i,j) + \begin{cases}
	 a^n_j (r_p - l_p), \hspace{1cm} i\in[1,p] \\
	 a^n_j  (r_{m\text{-}p} - l_{m\text{-}p}), \hspace{0.25cm} i \in[p,m-p] \\ 
	a^n_j  (r_i - l_i), \hspace{1cm} i  \in[m-p,m] 
\end{cases}
\label{Eq 1}
\end{equation}
\begin{equation}
L'(i,j)\\
 = L(i,j) + \begin{cases}
 	a^m_i (b_q - t_q), \hspace{1cm} i\in[1,q]\\
	a^m_i (b_j - t_j), \hspace{1cm} i \in[q,n-q]\\
	a^m_i (b_{n\text{-}q} - t_{n\text{-}q}),\hspace{.5cm}  i  \in[n-q,n] 
\end{cases}
\label{Eq 2}
\end{equation}
where $a^n_j = \tfrac{n-2j+1}{2(n-1)} $ and $l_i$ and $r_i$ denote the average intensity of respectively left and right edges of the local region of size $(2p \dotplus1) \times ( 2q \dotplus 1)$, to the $i_{th}$  row of the mask. Similarly, $t_j$ and $b_j$ denotes the average intensity of the upper and lower edges of the local region at the $j_{th}$ column of the mask.  Fig. 5(b) is showing the output of this method.
\subsubsection{De-hazing}
When rock material is dropped on the surface, is surrounded by scattering medium such as dust particles and water droplets, which may lead to decrease the robust detection of the object as shown in Fig 4(c). The removal of these effects are highly desired in the system which can be significantly increase the visibility of the objects. The performance of many vision algorithms e.g., feature detection, filtering, and photometric analysis will inevitably suffer from the biased and low-contrast scene, also provide the depth information and benefit many vision algorithms and advanced image editing \cite{5567108}. De-hazing function \cite{5567108} is used on the illuminated image to reduce the effects of water droplets and dust. The clear and correct position of the rock is estimated as shown in Fig. 4(d). 
\subsection{Dual foreground frame differencing approach}
\label{sec:2}
In this section, we obtain two foreground masks from frame differencing and \emph{BS-MOG} to estimate the static and moving objects.
\subsubsection{Frame Difference}
The first frame of the incoming video is initialized as current frame \texttt{Cf}, if the background image is empty then the \texttt{Cf} frame is stored as the background/reference image \texttt{Bi}. This background image will be same in the whole process. In this way, even if the \texttt{Cf} is changing at a fast pace, it will not affect the \texttt{Bi}. Subsequently, both \texttt{Cf} and \texttt{Bi} images are converted into \texttt{Lab} colour space and select the \texttt{L} luminance channel on calculate the absolute difference of corresponding pixels and stored as frame difference (\texttt{FdImage}) as described in Eq.~(\ref{Eq 3})
\begin{equation}
	\texttt{FdImage}_j = \texttt{Cf}_j - \texttt{Bi}  \hspace{1cm} j=0,1,2.....N 
\label{Eq 3}
\end{equation}
where $N$ is the total number of frames. In case of equality, when \texttt{Cf} and \texttt{Bi} are same then pixel intensities remain unchanged shown in first row of Fig.~5(c).
 \begin{figure}[H]
 \centering
 \includegraphics [width=0.45\textwidth ]{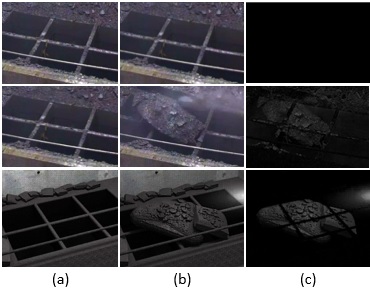}
\caption{Compute the frame difference: (a) Pre-processed background image \texttt{Bi}, (b) Current frame \texttt{Cf} (c) Difference image}
\label{fig:5}       
\end{figure}
Three different frames selected at different time intervals from synthetic video and computed the difference image \texttt{FdImage} as shown in Fig. 6 (a).
Foreground mask image \texttt{DFgMask} is obtained by applying 
gram based thresholding on the difference image \texttt{FdImage} as shown in Fig. 6(b). Thresholding takes difference image \texttt{FdImage} and classify all pixel values marked with “0” (background with black if the value is less than given threshold  \texttt{T}) and “1” (foreground with white if the pixel value is higher than \texttt{T}). 
Number of background pixels in the difference frame is higher than that of the foreground pixels, and the pixel value for the background is low compared to that of the foreground pixels as shown in Fig. 7. If we pick high threshold value the moving object pixels will be wiped out due to low number of peaks. Alternatively, low threshold value will generate false positive.
The important parameter in the thresholding process is the choice of the threshold value. The optimum value of \texttt{T} is picked by analysing the corresponding histogram of pixel values(global mean level from histogram) of the difference image of representative frame as described in \cite{Sengar2017MovingOD} and in such a way the peaks of background pixels eliminated.
 \begin{figure}[h]
 \includegraphics[width=80mm]{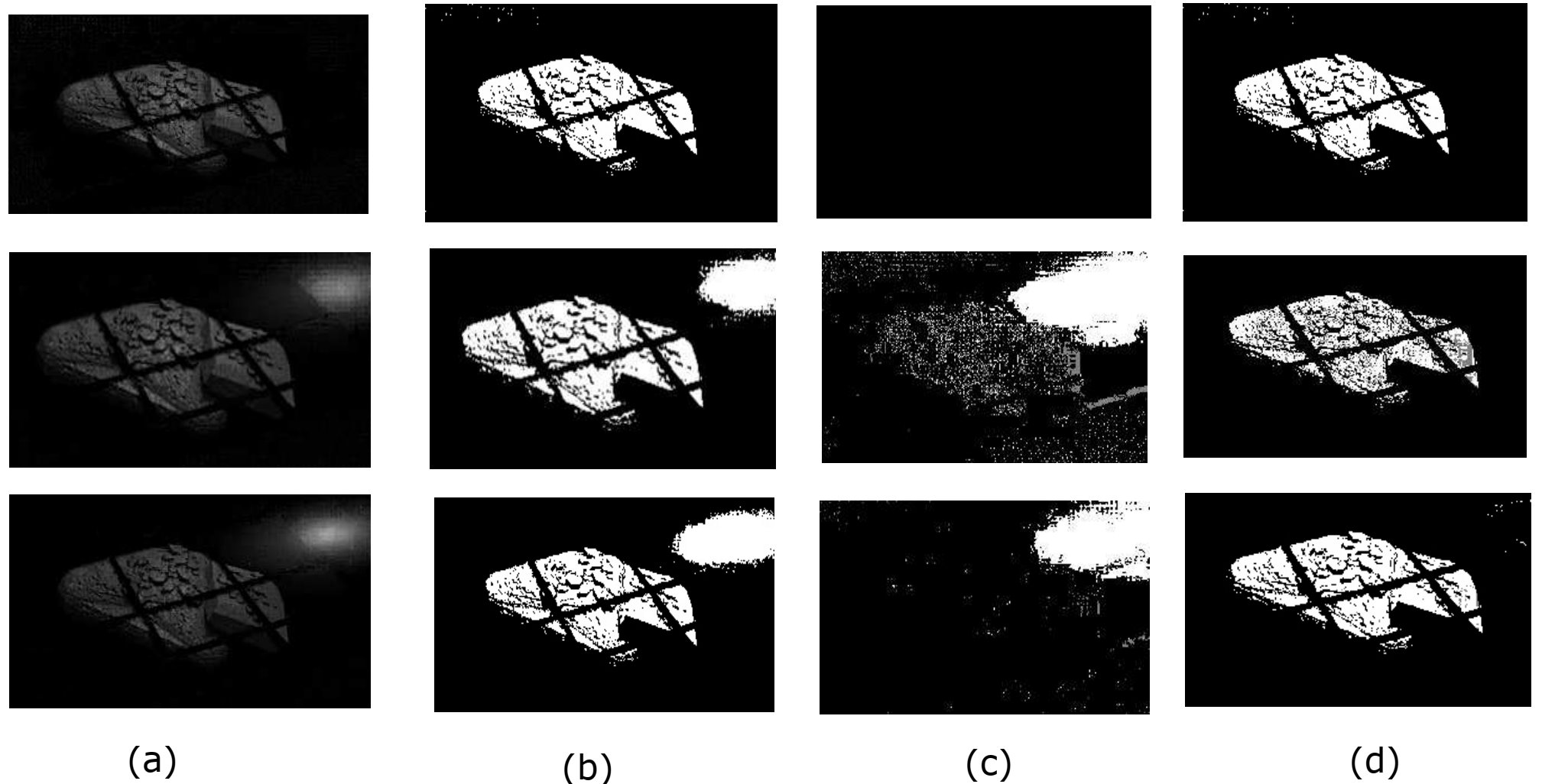}
\caption{Thresholding and foreground difference: (a)Difference image \texttt{FdImage} (b) Foreground image \texttt{DFgMask} (c) Foreground image \texttt{BS\_ FgMask} (d) foreground difference \texttt{FgMask}}
\label{fig:6}       
\end{figure}

Fig. 7 is showing the histogram of correspondent frames (rows) as shown in Fig. 6(a).There is a noticeable peak at low pixel values of the histogram due to the high rate of background pixel values in \texttt{FdImage} as compare to foreground pixels.
 \begin{figure}[h]
	\includegraphics[width=80mm]{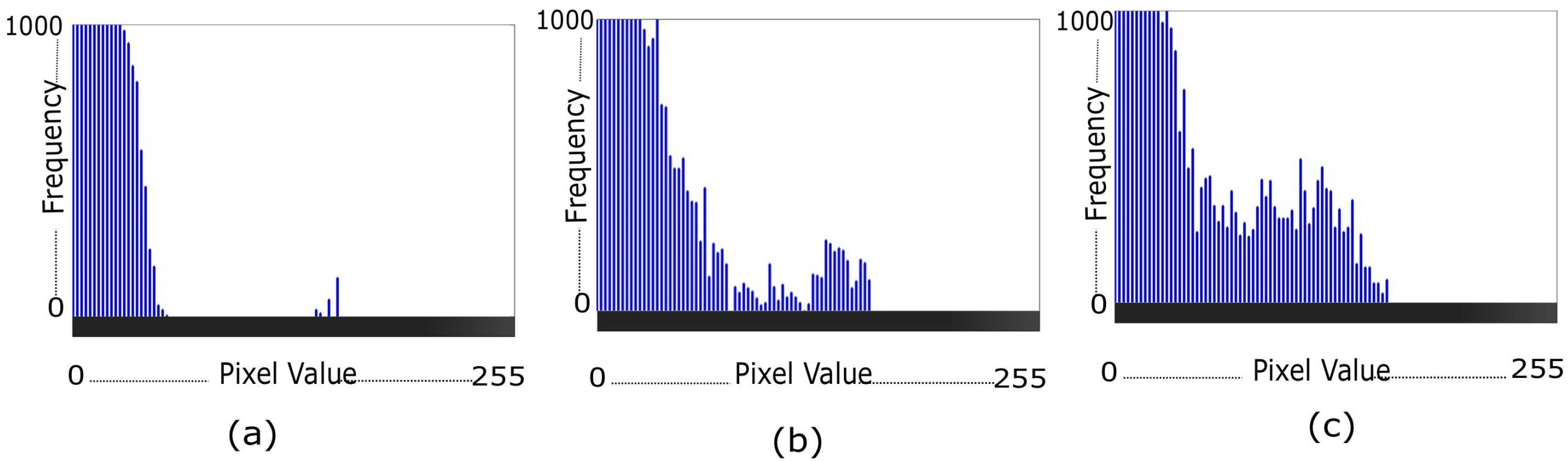}
	\caption{Histogram of pixel values of correspondent frame (Fig 6)}
	\label{fig:7}       
\end{figure}
\subsubsection{BS-MOG (Mixture of Gaussian)}
Background subtraction using \texttt{MOG} technique is used to detect the moving objects from static cameras. We used this method to detect the dust and moisture particles from the difference between the current and reference frame. As in the previous case, we compute the difference between current and static reference frame. In this method, the corresponding reference image is not constant, it changes over time and background pixel is modelled by a mixture of K Gaussian distributions (K = 3 to 5). The weights of the mixture represent the time proportions that those pixels stay in the scene. The probable background pixels have marked the ones, which stay longer and more static in the scene. Otherwise, it will be marked as foreground. Finally, dust and moisture particles pixel extracted with BS-MOG as shown in the Fig. 6(c).
\subsubsection{Static object detection}
At this stage, we have two binary foregrounds images from frame differencing \texttt{DFgMask} (Sect. 4.2.1) and MOG \texttt{BS\_ FgMask} (Sect. 4.2.2). Finally, foreground binary mask image \texttt{FgMask}  and estimated static object is detected by pixel wise classification of both foregrounds images as shown in Fig. 6(d).
\subsection{Post-processing and object detection}
Although pre-processing steps shows the significant elimination of noise but there are still obvious cavities left in \texttt{FgMask} binary image as shown Fig. 6(d). Therefore, post-processing steps: morphological operations and connected components analysis are applied on final binary foreground mask to suppress the noise for improving the final segmentation of rock piles. The eroded image is obtained by applying the \texttt{ELLIPSE} structuring element of size $3\times 3$ to eliminate thin constraints and then applied the dilation on the eroded version with same structure element but higher size \cite{Dougherty2003}.The pixel classification of rock was made on the dilated image by using the connected-components analysis to remove the isolated small size noisy blobs.  Once the connected components and their various properties like area, position and size have been generated, larger connected components were classified as desire object as shown in the Fig. 8 (b). Finally, draw the green line convex hull around the largest object as shown in Fig. 8(c).
\begin{figure}[!ht]
  \begin{center}
  \includegraphics[width=80mm]{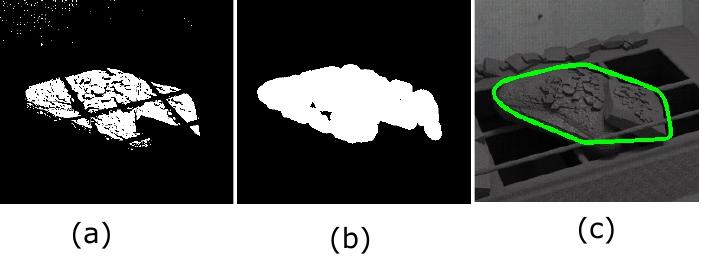}
\caption{Post-Processing: (a) Binary foreground image \texttt{FgMask}, (b) Post-Processing(c) Convex hull}
\end{center}
\label{fig:7}       
\end{figure}
\section{Experimental results and analysis}
In order to measure the performance of the system, four experiments are performed on datasets (Sec. 3). In real videos, we selected four frames (62, 200, 251, and 296) to show the results. As earlier pointed out, our main objective is to detect the static and moving objects in the \emph{ROI} and segmented the larger size of rock pile. Pre-processing is applied on \emph{ROI} image to reduce the effects of the poor contrast and lightening.  Histogram-based thresholding method is applied on original and pre-processed \emph{ROI} image. The estimation of rock is clearly observed after pre-processing as compare to the original image as shown in. Fig. 9(d).
\begin{figure}[!ht]
\begin{center}
     \includegraphics[width=80mm]{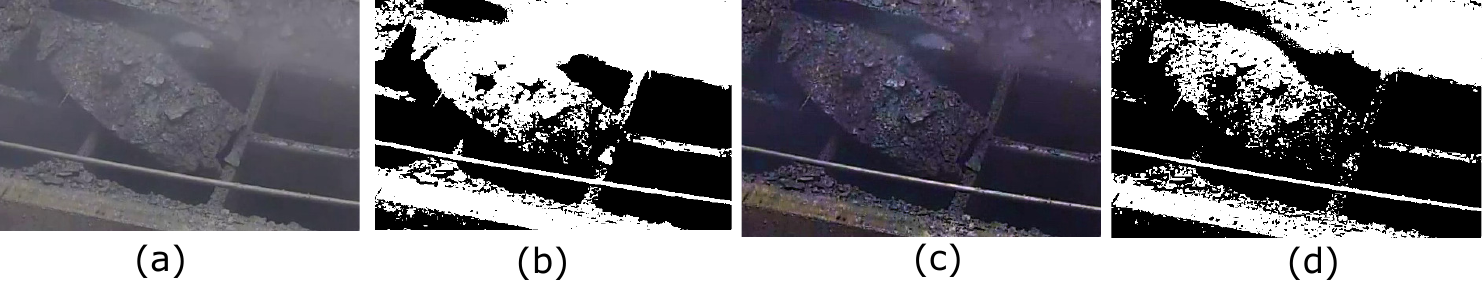}
\caption{Histogram based thresholding: (a) ROI image, (b) Thesholding ROI image,(c) Pre-processed ROI image, (d) Thresholding result on pre-processed image}
\label{fig:8}       
\end{center}
\end{figure}
Fig. 10 shows the results of selected frames from the right and left side cameras, where the first column is the \texttt{ROI} of current frame after pre-processing methods \texttt{Cf} and the second column is the frame differencing result with the static reference image. The third and fourth column presents the results of binary foreground images of frame differencing and \texttt{BS-MOG}.The next three columns show the subtracted foreground, post-processing and convex hull. 
In the second experiment, the proposed approach is applied on synthetic videos. Quantitative results achieved of larger rock piles detection and segmentation in synthetic videos from the right and left side camera with different shape and position as shown Fig.11.

To reduce the effects of dust material, water showering is applied in the station.  Some frames e.g. Fig.10 and Fig. 11 (2nd row of the right side camera), incorrect detection of rock is produced due to water showering. In this case, both camera and water showering have same direction and the intensity of water vapors pixels were strong and become the part of rock. To overwhelm this problem, the camera position should be different from water showering.  

\begin{figure}[!ht]
 \begin{center}
  \includegraphics[scale=.70]{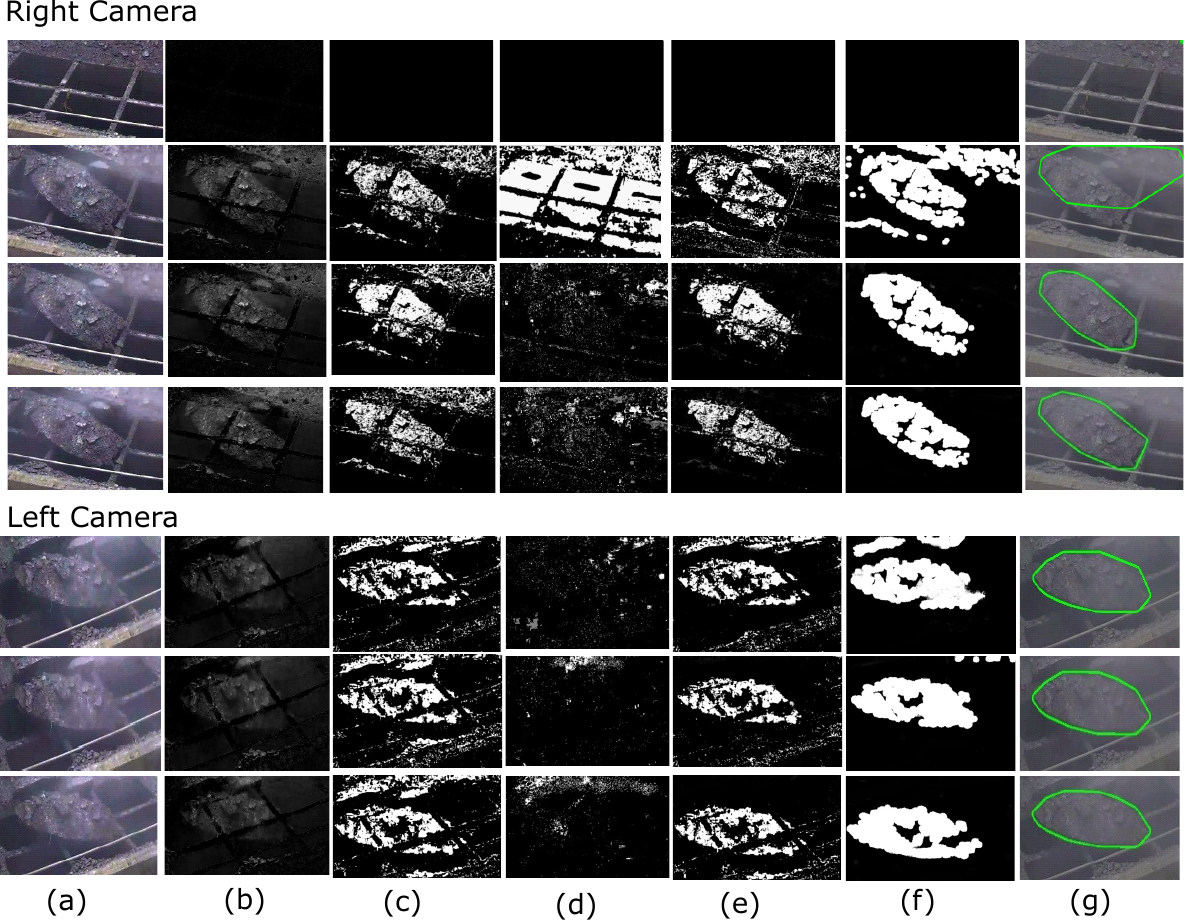}
\caption{Detection results of real videos (a) Current frame (62, 200, 251, and 296), (b) Frame difference result with static reference image, (c) Binary foreground image of b,(d)\texttt{BS-MOG} foreground mask (e)subtraction of moving pixels (f)Post-processing (g) final result with convex hull}
\label{fig:9}  
\end{center}     
\end{figure}
\begin{figure}[!ht]
 \begin{center}
   \includegraphics[width=0.45\textwidth]{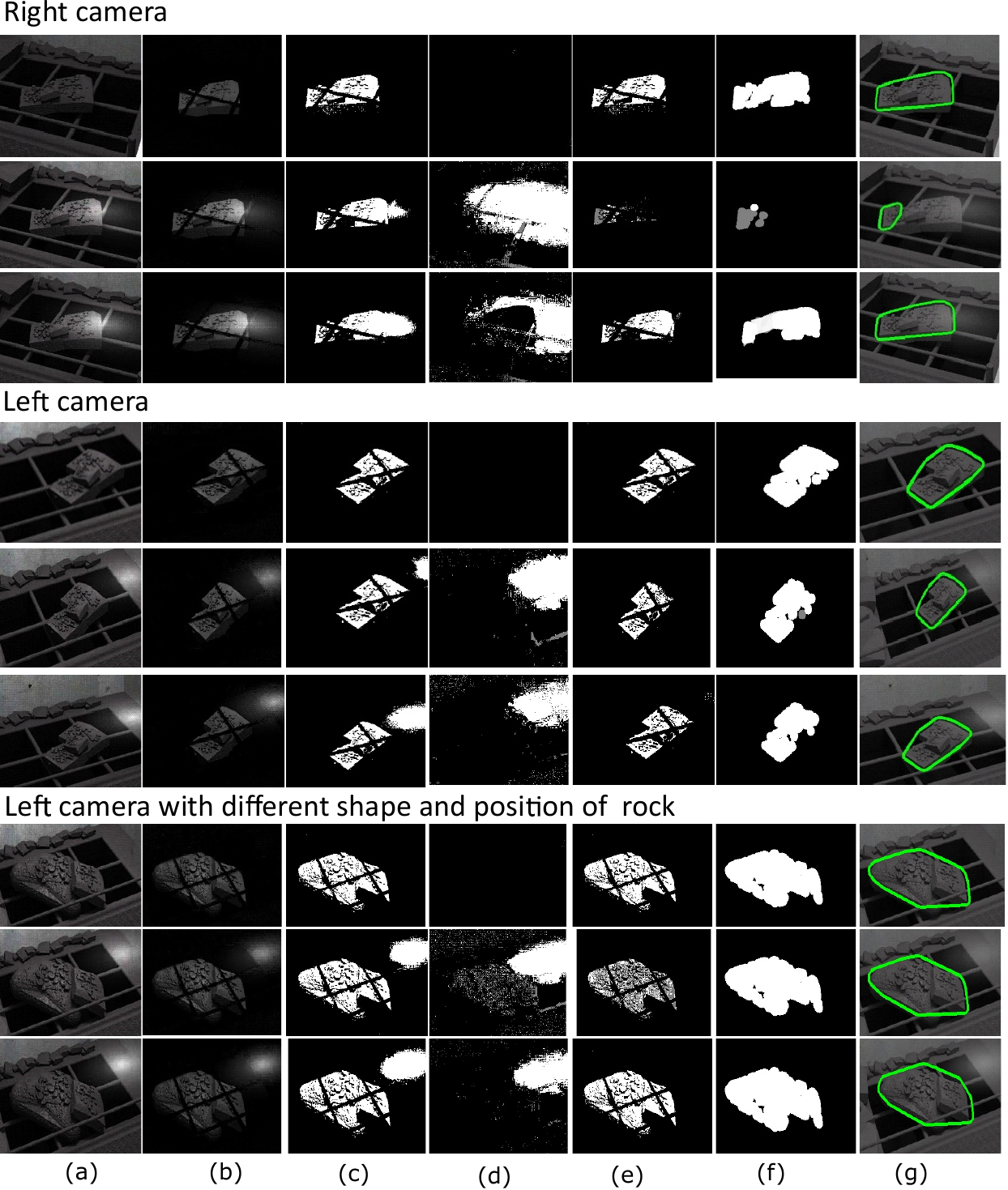}
\caption{Detection results of synthetic video: (a) Current frame (b)Frame difference result with static reference image,(c)Binary foreground image of b,(d)\texttt{BS-MOG} foreground mask (e)subtraction of moving pixels(f) Post-processing(g) final result with convex hull}
\label{fig:10}  
\end{center}     
\end{figure}
 
Furthermore, to test the system robustness we used two public datasets: AVSS 2007 (I-Lids) \cite{iLids} and CAVIAR \cite{CAVIAR}. Two different sequences AVSS-AB-Easy and PV-medium from AVSS 2007(I-Lids) are selected with image size: 720 x 576 pixels and 25 FPS. The main challenges are: detect the static object namely bag and illegally parked vehicles in the scenes. Static objects are successfully detected and segmented from the scenes as shown in green colour in Fig. 12 and 13. 

\begin{figure}[!ht]
 \begin{center}
   \includegraphics[width=0.45\textwidth]{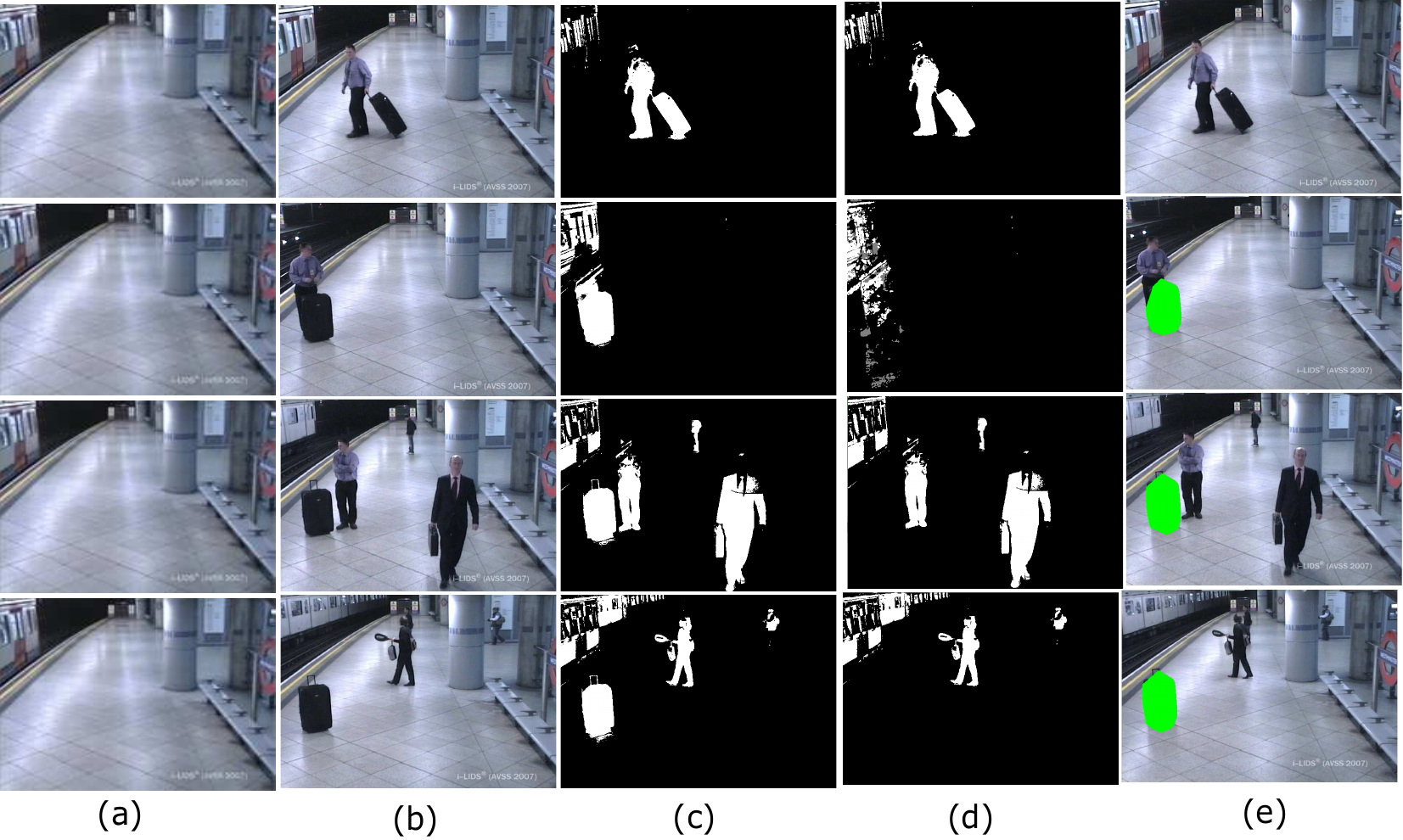}
\caption{Detection results of a bag in $AVSS-AB-Easy$ from i-Lids, (a) Background model (b) Current frame (1962, 2038, 2340, 3112), (c) Foreground mask \texttt{FdImage}, (d) Moving object detection \texttt{BS-MOG}, (e)Bag detection}
\label{fig:11}  
\end{center}     
\end{figure}
\begin{figure}[!ht]
 \begin{center}
   \includegraphics[width=0.50\textwidth]{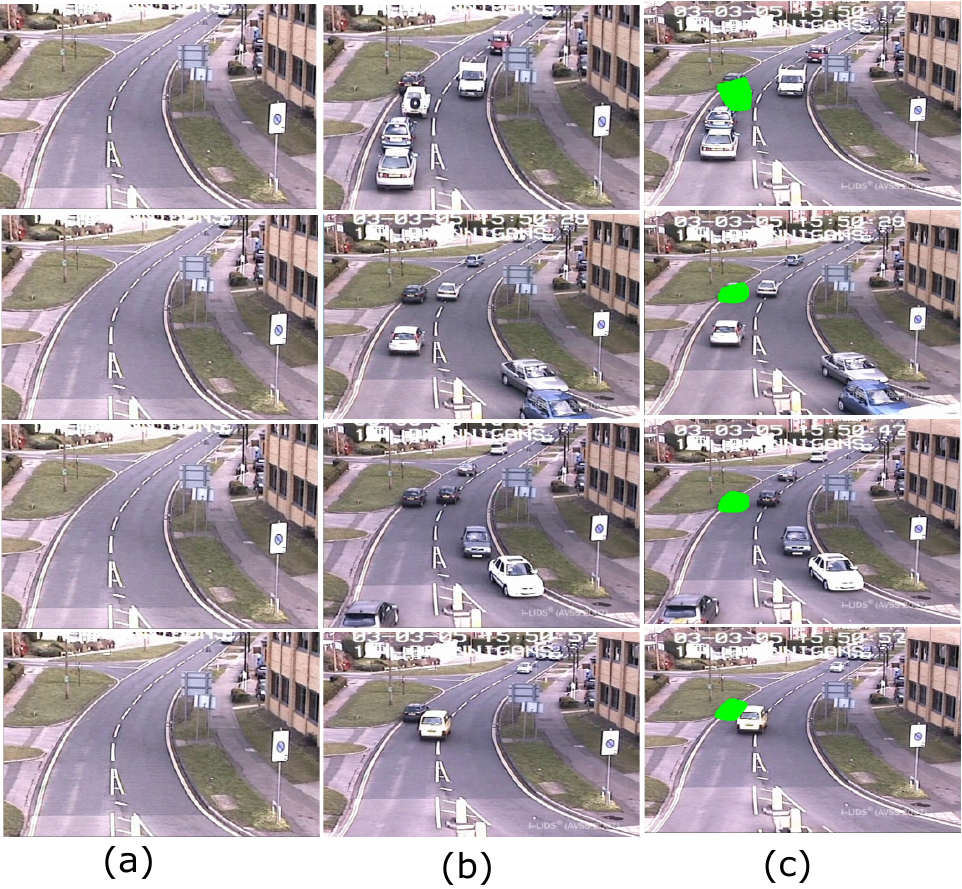}
\caption{Detection results of stop vehicle in $AVSS-PV-Medium$ sequence from i-Lids in dark shadows: (a) Background model, (b) Current frames (743,1050,1480,1745), (c) stopped vehicle detection}
\label{fig:12}  
\end{center}     
\end{figure}
The CAVIAR datasets comprise of different videos used in security surveillance applications (image size: 384 x 288 pixels, 25 FPS. We select the \texttt{LeftBag} video where a person carry a suitcase, leave it on the floor for short time. Suitcase was detected and marked as green colour as shown in Fig. 14.

\begin{figure}[!ht]
 \begin{center}
   \includegraphics[width=0.50\textwidth]{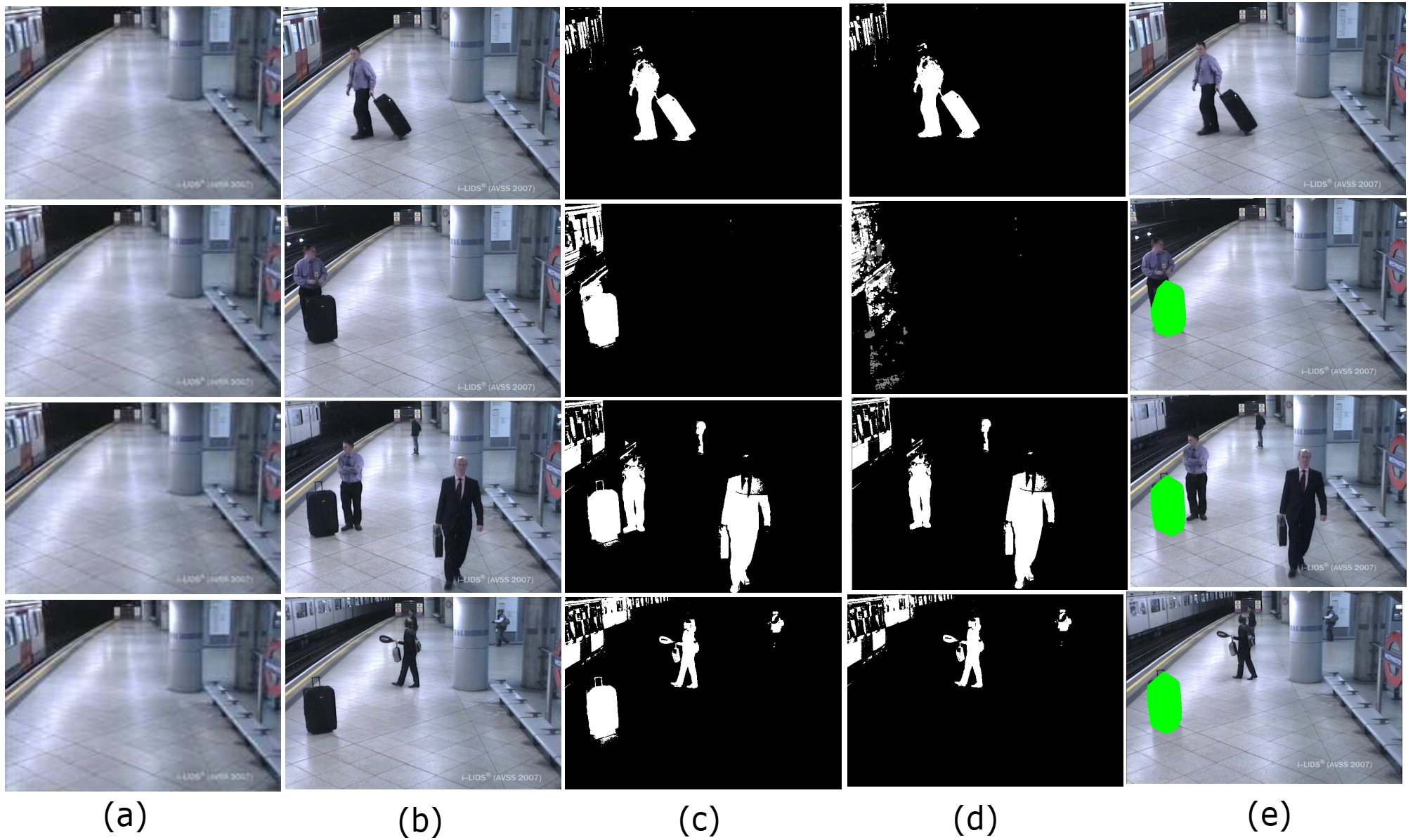}
\caption{Detection results of a bag in Leftbag sequence (courtesy of CAVIAR) : (a) Background model, (b) Current frames (743,1050,1480,1745), (c) Bag detection}
\label{fig:13}  
\end{center}     
\end{figure}
Results showed that static objects (rock, bag and vehicle) in different conditions were detected successfully. The first foreground frame differencing \texttt{DFgMask} approach with static reference image composes with pre-processing methods; detect both static and moving object pixels. The second foreground \texttt{BS\_ FgMask} detects only the moving particle pixels. The moving object information is obtained only in those frames where the object is in moving state e.g. rock dropping on the surface. The robust classification of static object pixels were achieved by subtracting moving object pixels \texttt{BS\_ FgMask} from frame \texttt{DFgMask}. Moreover, obstructions due to the surrounding moving objects do not cause the system to fail. The processing time is slightly varying due to pre and post-processing steps on all the frames.  Since pixel wise classifications used in proposed method, the processing time can be improved by using the parallel processing architectures e.g. by assigning each image pixels to a processor on the GPU.
\section{Conclusion}
\label{Sec:conclusion}
We present a frame differencing method for static object detection in different environmental conditions without any prior information about the object. At every scene, two foregrounds models are used for pixel classification. Static object information is achieved by subtracting the moving object pixels.  Proposed method takes full advantages of the \emph{BS} method and efficiently reduce the effect of noise from dust and moisture by applying the pre and post-processing steps.
The method was tested on real and synthetic video data sets and was found to be robust even if noise, illumination variations and scattered materials are present in the sequences. Furthermore, it was also successfully validated with two public data sets to detect the static objects.  Since the proposed method implements pixel wise classification and can be employed on parallel processors. As future work, we aim to extend the method to determine three-dimensional shape, position of the detected objects by making use of the multiple camera and segmentation of occluded objects.
\begin{acknowledgements}
This research was in part supported by the Kempe Foundations (SMK-1644.1).  The authors are thankful to Boliden AB for providing data and to Daniel Lindmark at Ume\aa\ University for assistance with producing the synthetic data. 
\end{acknowledgements}

\bibliographystyle{unsrt}

\end{document}